\pdfoutput=1
\documentclass[runningheads]{llncs}
\usepackage{epsfig}
\usepackage{graphicx}
\usepackage{amsmath}
\usepackage{amssymb}
\usepackage{mathtools} 
\usepackage{autobreak}

\usepackage{color}
\begin{document}
\title{Shared Latent Space of Font Shapes and Their Noisy Impressions}
%
%
\author{Jihun Kang\inst{1} \and
Daichi Haraguchi\inst{1} \and
Seiya Matsuda\inst{1} \and
Akisato Kimura\inst{2} \and
Seiichi Uchida\inst{1}\orcidID{0000-0001-8592-7566}}
\authorrunning{J. Kang et al.}
%
\institute{Kyushu University, Fukuoka, Japan
\email{\{jihun.kang\}@human.ait.kyushu-u.ac.jp}\and
NTT Communication Science Laboratories, NTT Corporation, Japan}
%
%
%
\maketitle              
\begin{abstract}
Styles of typefaces or fonts are often associated with specific impressions, such as heavy, contemporary, or elegant. 
This indicates that there are certain correlations between font shapes and their impressions. To understand the correlations, this paper realizes a shared latent space where a font and its impressions are embedded nearby. The difficulty is that the impression words attached to a font are often very noisy. This is because impression words are very subjective and diverse. More importantly, some impression words have no direct relevance to the font shapes and will disturb the realization of the shared latent space. We, therefore, use DeepSets for enhancing shape-relevant words and suppressing shape irrelevant words automatically while training the shared latent space. Quantitative and qualitative experimental results with a large-scale font-impression dataset demonstrate that the shared latent space by the proposed method describes the correlation appropriately, especially for the shape-relevant impression words. 
\end{abstract}
%
%
\section{Introduction}\label{sec:intro}
\begin{figure}[t]
    \centering
    \includegraphics[width=1.0\linewidth]{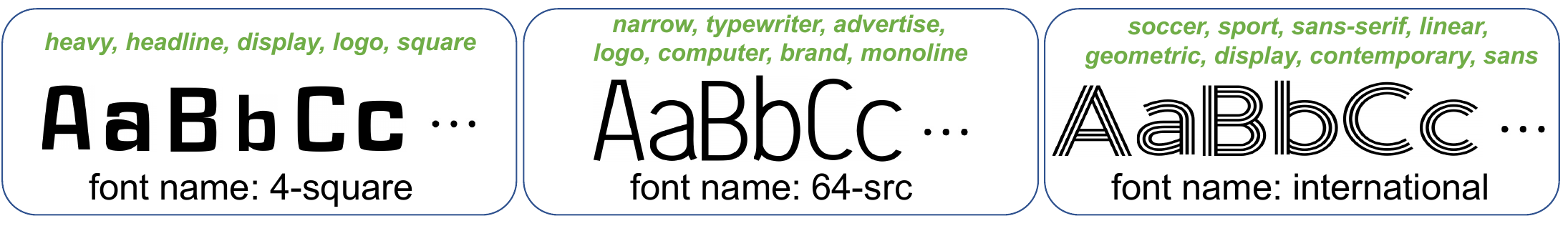}\\[-4mm]
    \caption{Examples of font images with their impression words (from \cite{chen2019large}.)}
    \label{fig:shape-and-impression}
    \bigskip
    \includegraphics[width=0.9\linewidth]{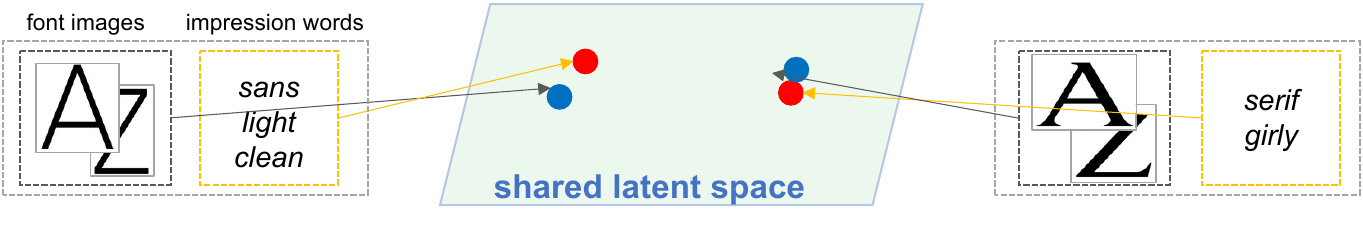}\\[-4mm]
    \caption{The shared latent space of font shapes and their impressions. The set of font images $\mathbf{X}_i$ and the set of impression words $\mathbf{W}_i$ of the $i$th font are embedded in the space by the embedding functions $f$ and $g$, respectively.}
    \label{fig:purpose}
    \bigskip
    \centering
    \includegraphics[width=1.0\linewidth]{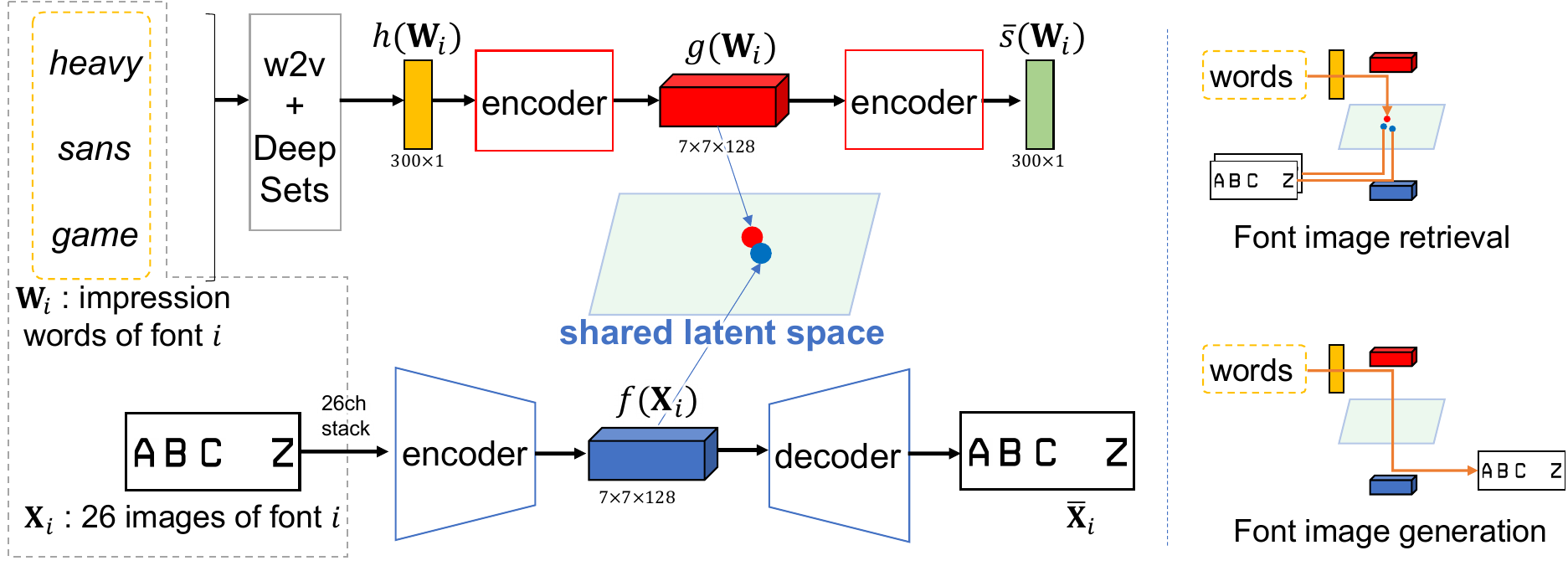}\\[-4mm]
    \caption{Left: an overview of the proposed method. Right: examples of applications.}
    \label{fig:overview}
\end{figure}

Font is multi-modal. This is because a font is comprised of not only a set of visible letter shapes (from `A' to `z') but only a set of impressions. Fig.~\ref{fig:shape-and-impression} shows several examples of fonts and their impressions from MyFonts dataset~\cite{chen2019large}.  For example, the font {\tt 4-square} is tagged with a set of impression words $\{$heavy, headline, display, logo, square$\}$. This is an interesting phenomenon in that a shape gives a specific impression; however, the correlation between shape and impression is not well studied in a reliable and objective data-driven analysis.\par

Our research aims to realize a {\em shared latent space} of the two modalities in order to understand their correlation. 
Fig.~\ref{fig:purpose} illustrates the shared latent space.  Let $\mathbf{X}_i$ denote the $i$-th font (i.e., a set of images from `A' to `Z' of the $i$-th font) and assume a set of $J_i$ impression words $\mathbf{W}_i=\{w_{i,1},\ldots,w_{i,j},\ldots,w_{i,J_i}\}$ are attached to the font $\mathbf{X}_i$. 
In the $d$-dimensional shared latent space, we expect that $f(\mathbf{X}_i)\sim g(\mathbf{W}_i)$ for all $i$, where the embedding functions $f$ and $g$ give $d$-dimensional representations of $\mathbf{X}_i$ and $\mathbf{W}_i$, respectively. Therefore, the realization of the latent space is the task of getting the representation functions $f$ and $g$ that satisfy this proximity condition.\par

\par
For realizing the shared latent space, we need to deal with the {\em noisiness} of the impression words. The noisiness comes from two reasons. First, the impression of a font is subjective and will be variable with its observers. The second and more serious reason is that impression words are often irrelevant to font shape.  In Fig.~\ref{fig:shape-and-impression}, an impression {\it soccer} is attached to the font {\tt international}. This impression might be attached because the font is used for the uniform of a soccer team. As revealed by this example, there are two types of impression words, {\em shape-relevant} and {\em shape-irrelevant}. The former type  (such as {\it sans-serif} and {\it heavy}) is our target; however, the latter will disturb the realization of the shared latent space and its effect should be weakened.
\par
\par
This paper proposes a novel method for realizing the shared latent space while weakening the effect of noisy (i.e., shape-irrelevant) impression words. Fig.\ref{fig:overview} shows the overall structure of the proposed method. It is a cross-modal embedding scheme and comprised of two autoencoders for word and image modalities. These autoencoders are co-trained so that $f(\mathbf{X}_i)\sim g(\mathbf{W}_i)$, while guaranteeing accurate reconstruction at each modality.  
Once we realize the shared latent space, it can be used for several applications, such as font image retrieval and font image generation, as shown in Fig.~\ref{fig:overview}; given a set of impression words, we can retrieve several existing font images or generate new font images. \par
A technical highlight of the proposed method is that it employs DeepSets~\cite{zaheer2017deep} for weakening shape-irrelevant impression words. Roughly speaking, DeepSets accepts a set as its input, converts each element into a feature vector internally, and finally outputs the average of feature vectors. If an element of the set is useless for a task, its feature vector will become close to a zero vector and thus its effect on the final output is minimized.  In our case, this set corresponds to a set of impression words, and the effect of the impression word that disturbs our task will be minimized. Note that DeepSets is also suitable to deal with the arbitrary number of impression words.\par 
%

%

Based on the above discussions, we can summarize the main technical contributions of this paper as follows:
\begin{itemize}
    \item This paper realizes a shared latent space for shape and impression by a novel cross-modal embedding scheme. To the authors' best knowledge, it is the first attempt to directly connect shapes (i.e., font images) and impressions by using a reliable large-scale dataset and machine-learning framework.
    \item Considering noisy impression words, we introduce DeepSets into the cross-modal embedding scheme. It also has another merit that we can deal with an arbitrary number of impression words for each font.
    \item Experimental analysis reveals the existence of two-type of impression words, shape-relevant, and shape-irrelevant. The former results in more correlated embedding in the shared latent space than the latter.
    \item Experimental results show that it is possible to retrieve and generate font images by specifying shape-relevant impression words.
\end{itemize}

\section{Related Work}\label{sec:related}
\subsection{Font shape and impression}

In the fields of psychology and marketing, the relationship between fonts and their impressions has been analyzed experimentally for many years~\cite{childers2002all,doyle2011mixed,lewis1989typographic,shaikl,Velasco2015}. These trials often use a small number of fonts. In fact, only 12 fonts are used in the rather recent trial~\cite{Velasco2015}. Nowadays, analysis with larger font image datasets~\cite{chen2019large,shinahara2019serif,odonovan,ikoma2020effect} has been conducted.  Among them, the font-impression dataset by O'Donovan et al.~\cite{odonovan} realizes impression-based font recommendation systems~\cite{choi2018fontmatcher,shirani2020let}. MyFonts dataset by Chen et al.~\cite{chen2019large} is a far more large dataset and used for impression-based font retrieval~\cite{chen2019large} and impression-specific font image generation~\cite{matsuda2021impressions2font}. 
%
\par
%

%
The recent attempts are rather application-oriented and thus do not focus on the essential correlation between font styles and the impressions. To the authors' best knowledge, this is the first attempt to understand the correlation between the shape (, or image) $\mathbf{X}_i$ and the impression words $\mathbf{W}_i$ of the $i$th font by embedding them into the same $d$-dimensional vector space to satisfy $f(\mathbf{X}_i)\sim g(\mathbf{W}_i)$ as possible, while weakening the effect of shape-irrelevant noisy impression words.\par

\subsection{Latent space embedding}
In multi-modal modeling of images and words (or texts), many attempts have been made for shared latent space embedding of the images and words. Socher et al.~\cite{socher2010connecting} have proposed a model that segments and annotates images by mapping images associated with the words to a latent semantic space.
The same group extended this idea~\cite{Socher2013} by incorporating a neural network-based representation learning scheme of the image modality. In this work, the word modality is encoded by a hand-crafted feature, and then the image modality is mapped to the fixed word modality.  
In the works focusing on neural language caption generation~\cite{fang2015captions,karpathy2015deep,vinyals2015show}, images and texts are not embedded into the same latent space explicitly, but image features by Convolutional Neural Networks (CNNs) are used as an input of Recurrent Neural Networks (RNNs) that generate textual information.\par
In the document analysis research field, Almaz\'an et al.~\cite{almazan} have published a pioneering work that a word image and its textual information are embedded into the same space for word spotting and recognition even in a zero-shot manner. Such an embedding strategy is nowadays extended to deal with a tough multi-modal task, called
Text VQA~\cite{biten2019scene}. Sumi et al.~\cite{sumi} realized a shared latent space between online and offline handwriting sample pairs and proved that a stroke order recovery is possible via the shared latent space.\par

\subsection{Representation learning for a set}
When each training sample comprises a different number of elements without any specific order, some machine learning architecture that accepts a set as its input sample is necessary. DeepSets~\cite{zaheer2017deep} has been proposed as a simple but powerful framework to deal with sets as samples. 
In recent years, Saito et al.~\cite{saitoexchangeable} have proposed the architecture to use sets by capturing the properties from the basis of set matching mathematically and have tried novel fashion item matching using sets.\par
In this paper, we treat the impression words $\mathbf{W}_i$ attached to the $i$th font as a set. The number of the attached words is different among fonts, as shown in Fig.~\ref{fig:shape-and-impression}. In addition, the words have no specific order. 
We, therefore, use DeepSets to treat  $\mathbf{W}_i$ as a set. Note that the other modality $\mathbf{X}_i$ is represented as a stack of images instead of a set because  $\mathbf{X}_i$ always contains a fixed number of elements from `A' to `Z.'


\section{MyFonts dataset~\cite{chen2019large}}
As the font dataset with impression words, we employ the dataset published by Chen et al.,~\cite{chen2019large}. This dataset, hereafter called the MyFonts dataset, comprises 18,815 fonts collected at {\tt MyFonts.com}. As shown in Fig.~\ref{fig:shape-and-impression}, each font is tagged with $0\sim 184$ impression words attached by crowd-sourcing. This means that the impression words have a large variability according to the crowd-sourcing workers' subjective bias. The vocabulary size of the impression words is 1,824. As noted in Section~\ref{sec:intro}, impression words are often shape-relevant (such as {\it heavy} in Fig.~\ref{fig:shape-and-impression}) but sometimes rather shape-irrelevant (such as {\it soccer}).
\par
Since we need to train several networks as a function $g(\cdot)$ with sufficient samples, we removed non-frequent impression words attached to less than 100 fonts. Consequently, we used 451 impression words in our experiments\footnote{It does not guarantee that each of the training and test sets contains more than 100 fonts for each of the 451 impression words.}. In addition, we removed the dingbat (pictorial symbols) fonts and the circled fonts from the MyFonts dataset by manual inspections by three persons. We also removed fonts without any impression words (after the above non-frequent word removal). 
Consequently, we used 9,980, 2,992, and 1,223 fonts for training, validating, and testing, respectively. Hereafter $\mathbf{\Omega}_{\rm train}$, $\mathbf{\Omega}_{\rm val}$
and  $\mathbf{\Omega}_{\rm test}$ denote the training, validation, and test font sets, respectively. We used 26 capital letter images of `A' to `Z' in each font in the later experiment since we found several fonts where small letter images are collapsed.
\par
\section{Shared Latent Space of Font Shape and Impression}
This section provides the method to train our model of Fig.~\ref{fig:overview}. 
The training is organized in a two-step manner for faster and more accurate convergence. We first perform two independent training pipelines as an initialization of the cross-modal embedding scheme. More specifically, we independently train two different autoencoders for font shapes (the bottom pipeline of Fig.~\ref{fig:overview}) and impression words (the upper pipeline). The latent vectors of those autoencoders correspond to $f(\mathbf{X}_i)$  and $g(\mathbf{W}_i)$, respectively. Second, the end-to-end co-training will be performed to embed those latent vectors into the shared space to satisfy the condition $f(\mathbf{X}_i)\sim g(\mathbf{W}_i)$, while keeping the autoencoders' outputs accurate enough.
\par
\subsection{Font shape encoding by autoencoder\label{sec:init-image}}
An autoencoder is used for generating the latent vector $f(\mathbf{X}_i)$ of the image modality, i.e., font shapes.
As shown at the bottom of Fig.~\ref{fig:overview}, the autoencoder accepts $\mathbf{X}_i$ as its input and
generate $\bar{\mathbf{X}}_i$ via an intermediate compressed representation $f(\mathbf{X}_i)$. Both of $\mathbf{X}_i$ and $\bar{\mathbf{X}}_i$ are 26 images (stacked as 26 channels) and expected to be similar with each other, i.e., $\mathbf{X}_i\sim \bar{\mathbf{X}}_i$, in order to guarantee that $f(\mathbf{X}_i)$ carries the original shape information of $\mathbf{X}_i$ sufficiently.
Note that $f(\mathbf{X}_i)$ is emitted from the autoencoder as a tensor of $7\times 7\times 128$, whereas it is flattened as a $d =7\times 7\times 128 = 6,272$-dimensional vector in the shared latent space. In the following, these two representations are not distinguished unless otherwise mentioned ($g(\mathbf{W}_i)$ also).
\par
The encoder ($\mathbf{X}_i \mapsto f(\mathbf{X}_i)$) is based on ResNet18 (pre-trained with ImageNet) and the decoder ($f(\mathbf{X}_i) \mapsto \bar{\mathbf{X}}_i$) is comprised of several deconvolutional layers. (See Section~\ref{sec:implementation-detail} for the detail.)
They are trained to minimize the construction loss function $L_{\rm shape}= \sum_{i=1}^N{\left\|\mathbf{X}_i - \bar{\mathbf{X}}_i\right\|}$,  where $N$ is the number of fonts used for training.
\par
%
\begin{figure}[t]
    \centering
    \includegraphics[width=0.75\linewidth]{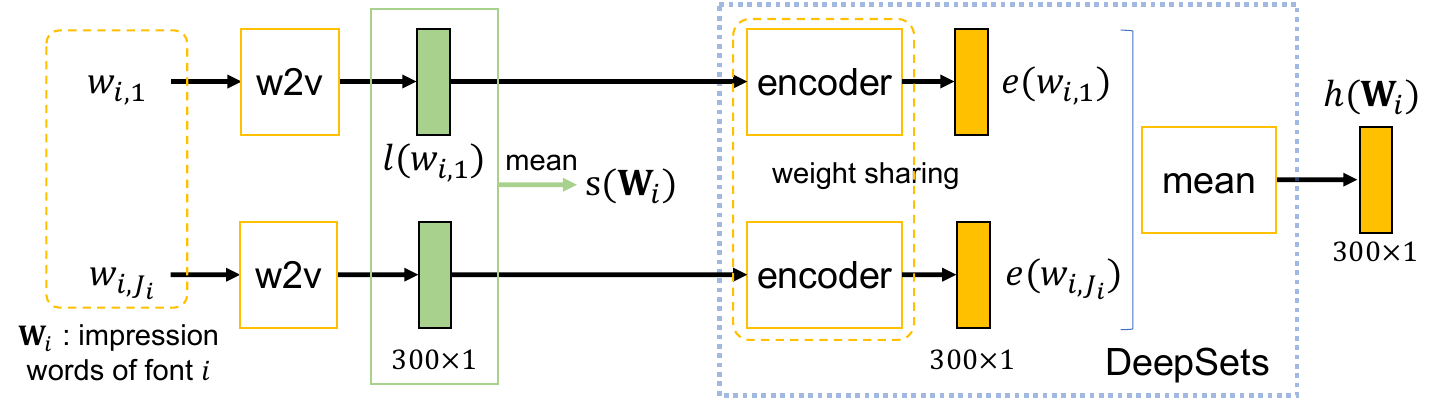}\\[-3mm]
    \caption{Word vector by aggregating the word vectors of multiple impression words by word2vec (w2v) and DeepSets.}
    \label{fig:deepset}
\end{figure}

\subsection{Noise-tolerant impression word encoding by DeepSets\label{sec:init-word}}
Like the image modality, an autoencoder is used for generating the impression word vectors $g(\mathbf{W}_i)$. However, the word modality requires extra modules to accept an arbitrary number of impression words $\mathbf{W}_i=\{w_{i,1},\ldots,w_{i,j},\ldots,\allowbreak w_{i,J_i}\}$ as its input.
Moreover, each impression word $w_{i,j}$ should be converted to a semantic vector so that similar impression words give similar effects to the system.
Therefore, as shown in Fig.~\ref{fig:deepset},  the impression word $w_{i,j}$ is converted to a semantic vector $l(w_{i,j})$ by Word2vec~\cite{mikolov2013distributed} (pretrained by Google News dataset), and then all the semantic vectors are aggregated to 
 a single fixed-dimensional vector $h(\mathbf{W}_i)$ by DeepSets~\cite{zaheer2017deep}. \par
%
%
Fig.~\ref{fig:deepset} shows how DeepSets converts the $J_i$ semantic vectors $\{l(w_{i,j}) | j=1,\ldots,J_i\}$ into a single vector $h(\mathbf{W}_i)$. DeepSets has two functions: a trainable encoding 
scheme, or representation learning, and an aggregation scheme. The former is a deep neural network and gives a new representation $e(w_{i,j})$
for the word2vec vector $l(w_{i,j})$. The latter is the simple averaging process
$h(\mathbf{W}_i) = \sum_j e(w_{i,j})/J_i$. This simple aggregation scheme allows accepting any number of impression words.
\par
The most promising property of DeepSets for our task is that it can automatically learn the feasibility of impression words for realizing the shared latent space. Therefore, if an impression word $w_{i,j}$ is shape-irrelevant and disturbs the realization, its effect will be weakened. As the result, even though the relationship $f(\mathbf{X}_i) \sim g(w_{i,j})$ will not hold for the shape-irrelevant word $w_{i,j}$,
the relationship will still hold for most shape-relevant words.
\par
As shown at the top of Fig.~\ref{fig:overview}, the autoencoder for the impression word modality 
accepts $h(\mathbf{W}_i)$ as its input and derives the latent representation $g(\mathbf{W}_i)$.
Note that if we train the autoencoder and DeepSets in an end-to-end manner to minimize the reconstruction loss of $h(\mathbf{W}_i)$, it results in the trivial solution that $h(\mathbf{W}_i)= \bar{h}(\mathbf{W}_i)= 0$. We, therefore, train them to minimize the loss function $L_{\rm impression} = \sum_{i=1}^N\| s(\mathbf{W}_i) - \bar{s}(\mathbf{W}_i)\|$, where $s(\mathbf{W}_i)=\sum_j l(w_{i,j})/J_i$ (as shown in Fig.~\ref{fig:deepset}) and $\bar{s}(\mathbf{W}_i)$ is the decoder output.

\subsection{Co-training for the shared latent space\label{sec:co-training}}
After the pre-training of the autoencoder for both modalities, all the modules of both modalities are co-trained to realize the shared latent space. From its purpose to have $f(\mathbf{X}_i) \sim g(\mathbf{W}_i)$, we have the  loss function $L_{\rm share} = \sum_{i=1}^N\| f(\mathbf{X}_i) - g(\mathbf{W}_i) \|$.
Consequently, the overall loss function of the proposed method becomes 
$L = L_{\rm shape} + L_{\rm impression} + L_{\rm share}$.
In the process of minimizing the loss function $L$, the weights of all autoencoders and DeepSets are trained simultaneously. During this, we expect that the effect of the shape-irrelevant impression words that have no clear correlation with font shapes will be minimized in DeepSets.\par

\section{Experimental Results}\label{sec:results}
\subsection{Implementation details\label{sec:implementation-detail}}
For the image modality, the encoder ($\mathbf{X}_i \mapsto f(\mathbf{X}_i)$) is ResNet18 (pre-trained by ImageNet) that have additional convolution layer at the last whose kernel size is $1\times1$ and number of channel is 128. The decoder ($f(\mathbf{X}_i) \mapsto \bar{\mathbf{X}}_i$) is $\rm D_{(512,1,0)}^{1 \times 1}$--$\rm R$--$\rm D_{(256,2,1)}^{4 \times 4}$--$\rm R$--$\rm D_{(128,2,1)}^{4 \times 4}$--$\rm R$--$\rm D_{(64,2,1)}^{4 \times 4}$--$\rm R$--$\rm D_{(32,2,1)}^{4 \times 4}$--$\rm R$--$\rm D_{(26,2,1)}^{4 \times 4}$ where $\rm D$ and $\rm R$ show a deconvolution layer and a ReLU function respectively. The parenthesized description shows (channels, stride, padding) and the superscript shows the kernel size. For the impression word modality, the encoder ($h(\mathbf{W}_i) \mapsto g(\mathbf{W}_i)$) is $\rm F_{(1024)}$--R--$\rm F_{(2048)}$--R--$\rm F_{(6272)}$, and the decoder ($g(\mathbf{W}_i) \mapsto \bar{s}(\mathbf{W}_i)$) is $\rm F_{(2048)}$--R--$\rm F_{(1024)}$--$\rm R$--$\rm F_{(300)}$ where $\rm F$ shows a fully-connected layer.
Note that the parenthesized description shows hidden units.
\par
The entire network is trained by the training font set $\mathbf{\Omega}_{\rm train}$ (9,980 fonts) and tested by the test set
$\mathbf{\Omega}_{\rm test}$ (1,223 fonts). The hyper-parameters and the training epochs are optimized by  $\mathbf{\Omega}_{\rm val}$ (2,992 fonts).\par

\begin{figure}[t]
    \centering
    \includegraphics[width=0.75\linewidth]{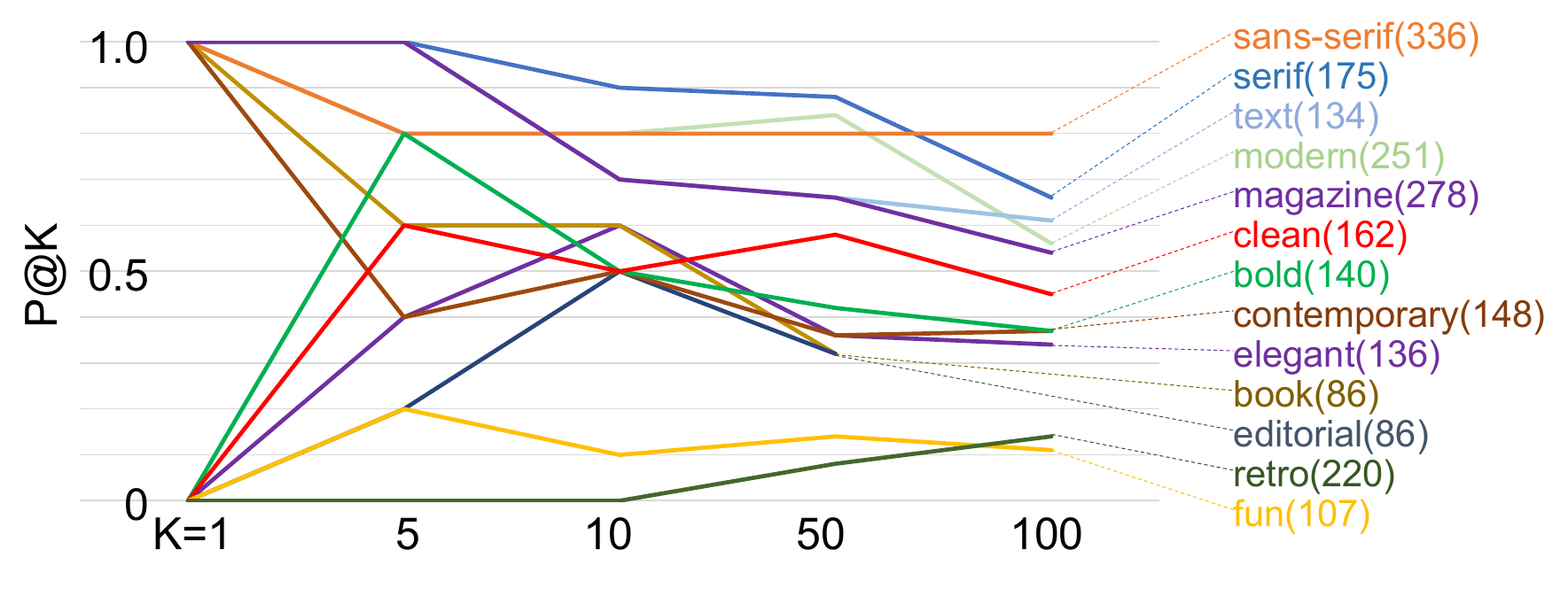}\\[-5mm]
    \caption{P@K($\uparrow$) for several impression words. The parenthesized number is the number of test fonts having the impression word.}
    \label{fig:P@10}
\end{figure}
\subsection{Quantitative analysis}

We have conducted font image retrieval from a given set of impression words, which is an application task shown in Fig.~\ref{fig:overview}.
If a font shape $\mathbf{X}_i$ and its corresponding impression words $\mathbf{W}_i$ are embedded appropriately while satisfying  $f(\mathbf{X}_i)\sim g(\mathbf{W}_i)$, we can retrieve the font $\mathbf{X}_i$ from $\mathbf{W}_i$ by a simple nearest neighbor 
search in the shared latent space. In the following, we use a simpler setup that uses only a single impression word as the query for the retrieval. This setup allows us to understand how individual impression words are more shape-relevant or irrelevant. \par
Fig.~\ref{fig:P@10} shows the quantitative retrieval performance  on the test set $\mathbf{\Omega}_{\rm test}$. The performance is 
measured by precision at $K$ (P@$K$) at different $K$. P@$K$ indicates 
the ratio of the correct fonts among $K$ retrieved fonts. More specifically, we first retrieve the $K$ nearest fonts for the specified impression word $w$ by the nearest neighbor search in the latent space.  Therefore, each retrieved font $X$ will satisfy $f(\mathbf{X})\sim g(w)$. Then, if a font $X$ has the impression word $w$ in its tag set, it is a correctly retrieved font. If P@$K$ is 1, all the $K$-neighboring font shapes have the impression word $w$ in their ground-truth. \par
Fig.~\ref{fig:P@10} shows P@$K$ for 13 impression words which are 11 words with the highest P@10 and two words, {\it retro} and {\it fun}, with rather lower P@10 values (0 and 0.1, respectively). Most of the 11 words with higher P@10 are obviously shape-relevant words, such as {\it serif}, {\it sans-serif}, and {\it bold}. This proves that our framework can successfully learn the representation describing the relationship between font shapes and their impressions. It is also noteworthy that more subjective impression words such as {\it modern} and {\it elegant},  have a high P@10 value. 
Although the ``elegant''ness of a font may vary among people, this result indicates there are common shape-relevant characteristics about it. 
\par

\begin{table}[t]
 \caption{Average retrieval rank, where (*) indicates a similar setup to \cite{Socher2013}.}
 \label{table:list}\vspace{-2mm}
 \centering
  \begin{tabular}{l|r|r}
    \hline
        Method & $R_{\rm{image}\to\rm{word}}$ & $R_{\rm{word}\to\rm{image}}$\\ \hline\hline
        Independent & 608.7 & 612.5\\ 
        Image $\mapsto$ Word(*) & 608.1 & 612.2\\
        Word $\mapsto$ Image(*) & 516.9 & 553.0\\
        Proposed & {\bf 172.6} & {\bf 356.6}\\ \hline
  \end{tabular}
  \vspace{-3mm}
\end{table}

Table~\ref{table:list} shows a more overall evaluation result of font image retrieval performance. By giving an impression word set $\mathbf{W}_i$ of the $i$-th test font as a query, all the images $\mathbf{X}\in\mathbf{\Omega}_{\rm test}$ are then ranked by the distance $\|f(\mathbf{X})-g(\mathbf{W}_i)\|$. Then, the rank $r_i$ of the correct image $\mathbf{X}_i$ among $|\mathbf{\Omega}_{\rm test}|$ images is determined. Finally, the average retrieval rank $R_{\rm{word}\to\rm{image}}=\sum{r_i}/|\mathbf{\Omega}_{\rm test}|$ is the evaluation metric in this evaluation. 
In a similar manner, we can obtain the average rank $R_{\rm{image}\to\rm{word}}$ for the task of word retrieval with a given font image. \par
To our best knowledge, this is the first attempt at the cross-modal embedding of font impression and font shape into the shared latent space;
therefore, there is no appropriate comparative baseline for this study. 
We, therefore, consider the following ablation cases to confirm the advantage of the proposed method.
\begin{itemize}
    \item Proposed: $f(\mathbf{X}_i)$ and $g(\mathbf{W}_i)$ are embedded by the co-trained encoder (Section~\ref{sec:co-training}).
    \item Independent: $f(\mathbf{X}_i)$ and $g(\mathbf{W}_i)$ are embedded by the encoders by the initial training steps (Sections~\ref{sec:init-image} and \ref{sec:init-word}). No co-training has been made.
    \item Image $\mapsto$ Word: After initializing the encoders, the encoder for the word modality is fixed. The encoder and decoder of the image modality are then trained so that $f(\mathbf{X}_i)\sim g(\mathbf{W}_i)$. This setup is very similar to a well-known previous research named cross-modal transfer~\cite{Socher2013}.
    \item Word $\mapsto$ Image: The opposite of the Image $\mapsto$ word setting. After initializing the encoder for the image modality, the encoder for the image modality is fixed. The encoder and decoder of the word modality are then trained so that $f(\mathbf{X}_i)\sim g(\mathbf{W}_i)$.
\end{itemize}
\par
Table~\ref{table:list} indicates that the proposed method greatly outperformed the others. 
It also indicates that both Image $\mapsto$ Word and Word $\mapsto$ Image settings did not work well, which implies that our proposal with joint optimization of cross-modal autoencoders was a key for obtaining meaningful representations.

\par
%



\begin{figure}[!t]
    \centering
\begin{minipage}[t]{0.48\linewidth}
    \centering
    \includegraphics[width=0.985\linewidth]{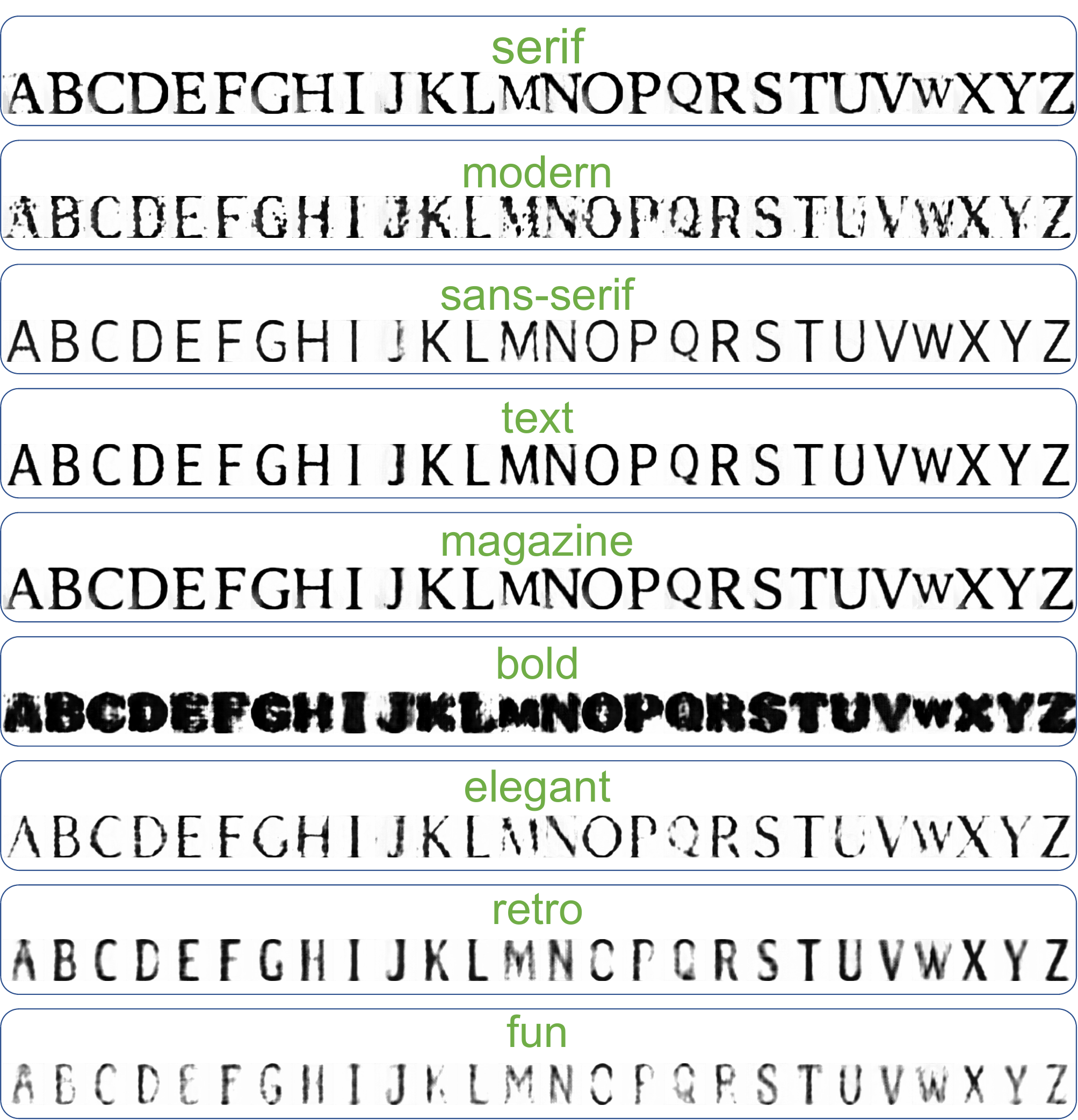}\\(a)~Single impression word
 \end{minipage}
 \begin{minipage}[t]{0.48\linewidth}
     \centering
     \includegraphics[width=1\linewidth]{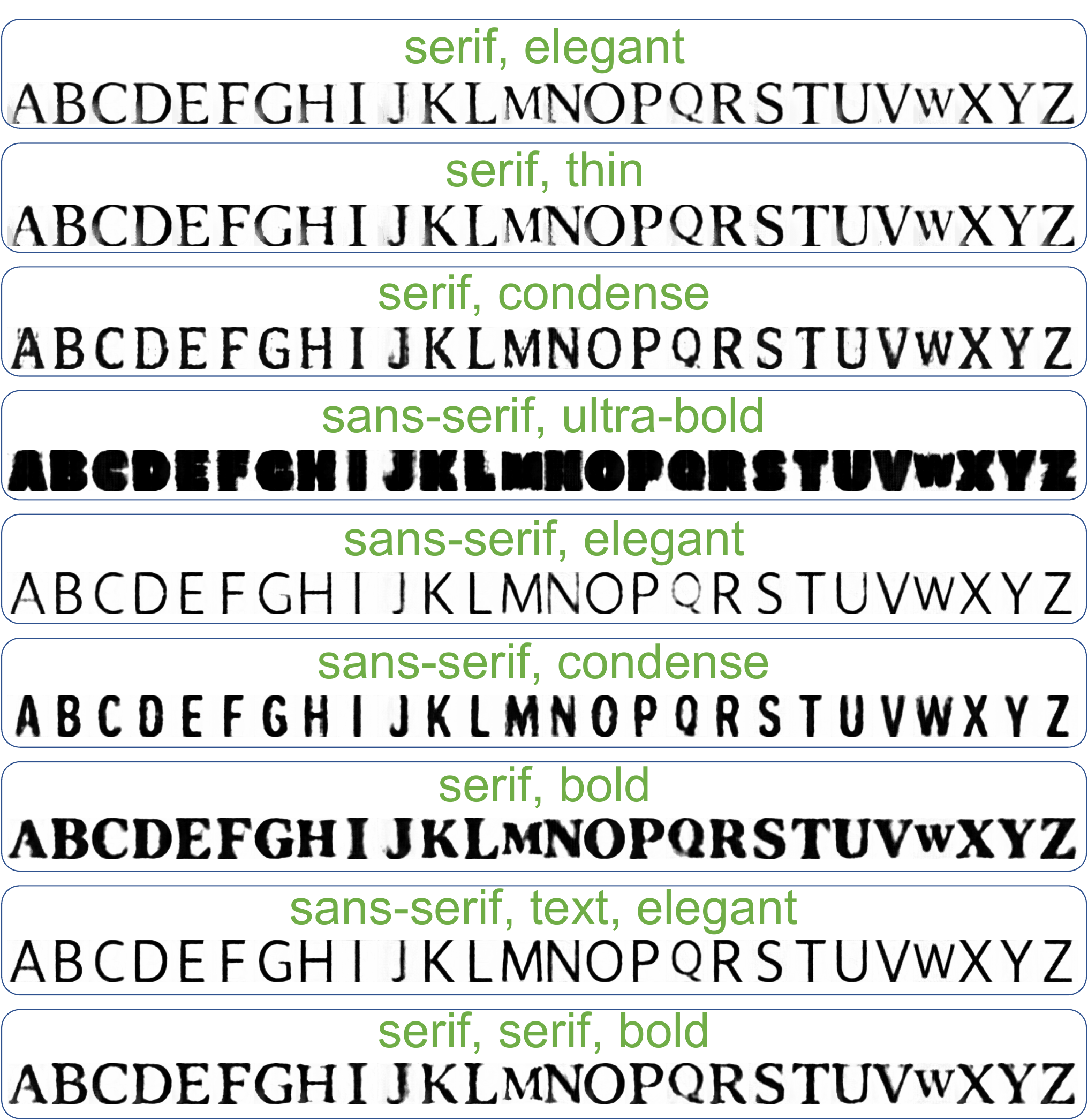}\\(b)~Multiple impression words
 \end{minipage}\medskip\\[-3mm]
 \caption{Generating images from a single impression word and multiple impression words. Note that {\it retro} and {\it fun} are shape-irrelevant words with lower P@K, whereas the others are shape-relevant.}
    \label{fig:imp_to_font}
\end{figure}

\par

\subsection{Font image generation with specific impressions}

As shown in Fig.~\ref{fig:overview}, we can use the shared latent space for generating font images with specific impressions. 
Feeding $g(\mathbf{W}_i)$ to the decoder of the image modality, we can generate the alphabet images from `A' to `Z' in a stacked manner.  
This is not only an interesting application but also a test to understand how the cross modal embedding is successful for each impression word. 
If a generated font image for a word $w$ is not appropriate, it indicates that $w$ is a shape-irrelevant word and thus $f(\mathbf{X})\not\sim g(w)$.
\par

Fig.~\ref{fig:imp_to_font}(a) shows the results when generating font images with a single impression word. 
For the shape-relevant words such as {\it serif} to {\it sans-serif}, legible font images with the specified impression are generated successfully, thanks to the property $f(\mathbf{X})\sim g(w)$. 
We obtained the expected results also for shape-irrelevant words such as {\it fun} and {\it retro}. Since there is no specific trend in the shapes for shape-irrelevant words, the generated font images are in a neutral style. 
Fig.~\ref{fig:imp_to_font}(b) shows the results when generating font images with multiple impression words. 
The font images generated by specifying two shape-relevant words such as ({\it serif}, {\it thin}) and ({\it sans-serif} and {\it bold}) become a mixed style successfully. 
The last example shows the case that the same word is specified twice; according to the nature of DeepSets, we can strengthen an impression by this strategy.\par
\begin{figure}[!t]
    \centering
    \centering
    \includegraphics[width=0.8\linewidth]{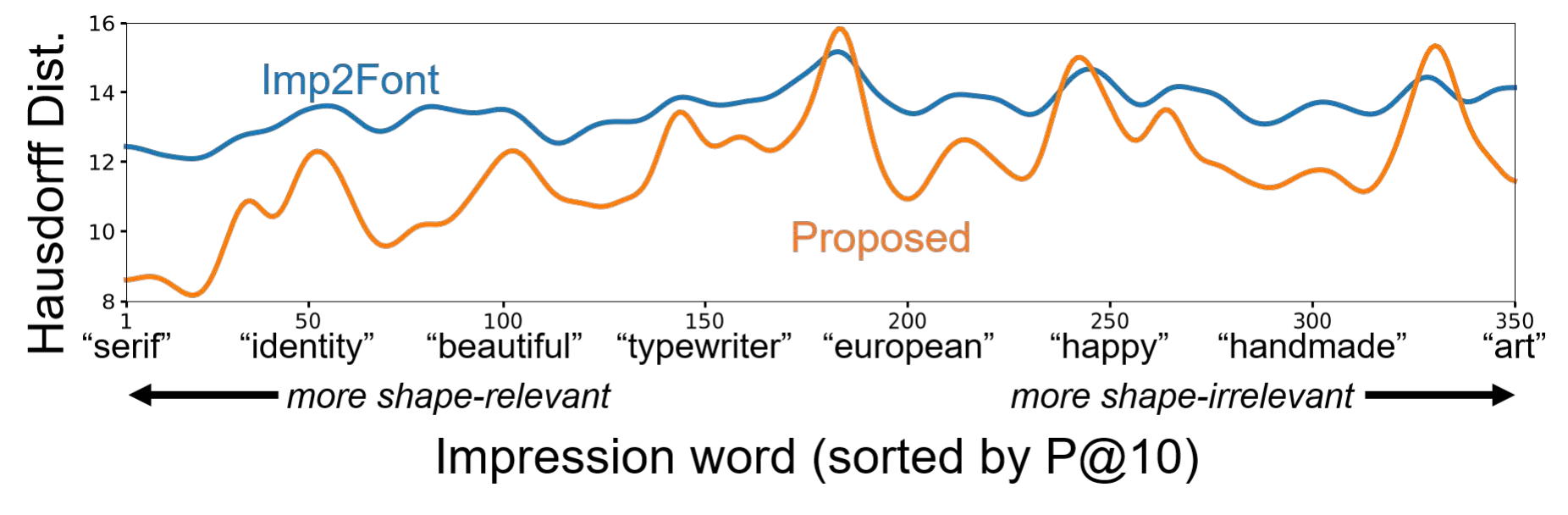}\\[-4mm]
 \caption{Quality of the generated font image for each impression word. The quality is measured by Hausdorff distance ($\downarrow$). The words are sorted by the P@10 ranking, i.e.,  shape-relevance. }
    \label{fig:distance}
\end{figure}
We evaluated the quality of generated font images quantitatively by Hausdorff distance. We compared the proposed method with the Impressions2Font~\cite{matsuda2021impressions2font}, which is a GAN-based method for generating font images from impression words. 
In this experiment, we generate font images from each impression word $w$ of a test font $\mathbf{X}$ by using the proposed method and Impression2Font, respectively\footnote{Precisely speaking, 34 compounded impression words, such as {\it caps-only} and {\it t-shirt}, are not acceptable by Impression2Font and thus omitted in the evaluation. Moreover, since we use P@10 in the evaluation of Fig.~\ref{fig:distance}, we also remove the impression words attached only to less than 10 test fonts. Consequently, we used 350 impression words in this experiment.}, and then compare the generated image with $\mathbf{X}$. For the comparison by Hausdorff distance, the generated images are binarized by the Otsu method and then converted to edge images by the Canny method. The Hausdorff distance is calculated at each of 26 alphabets and then median over them\footnote{Impression2Font is GAN and thus can generate different images with different random value inputs. Therefore, we used 10 random value inputs sampled from a standard normal distribution and generate $10\times 26$ images. Consequently, we use the median of all 260 Hausdorff distance values.}.
\par
Fig~\ref{fig:distance} shows the experimental results, where the horizontal axis corresponds to the impression word $w$ sorted by the P@10 ranking for the image retrieval task\footnote{Since the original transitions of the Hausdorff distance values show more jaggedness that hides their general trends, we applied a smoothing filter to the original transitions for getting the curves of Fig~\ref{fig:distance}.}. 
The results demonstrate the effectiveness of the proposed method compared with Impression2Font, especially for impression words with high P@10 rankings. This means the proposed method could generate font images more similar to the ground-truth images for shape-relevant impression words $w$.
As the ranking of P@$K$ decreases, the distance by the proposed method gradually increases, which implies that shape-irrelevant impression words with a weak relationship between impression and shape are not embedded in the shared space. In other words, this result simply reflects the fact that it is difficult to estimate the font shape from shape-irrelevant words.


\section{Conclusion}\label{sec:conclusion}
This paper showed that it is possible to realize a shared latent space where a font shape image and its multiple impression words are embedded as similar vectors. Through the shared latent space, we can handle font shapes and their impressions in a unified manner, which can lead us to generate and retrieve font images with specific fonts. Technically, we need to deal with shape-irrelevant impression words because they might disturb the unification; for this purpose, we incorporate DeepSets that can automatically weaken their effect.
Experimental results revealed the existence of shape-relevant and shape-irrelevant impression words. The shape-relevant words give a higher correlation with their corresponding font shapes. The experimental results also show the possibility of impression-specific font retrieval and font generation by specifying shape-relevant impressions.
\par
Future work will focus on additional experiments of font impression evaluation by translating font images to impression words via the shared latent space. We are also planning to standardize the impression words based on their degree of shape-relevance. \par

\bibliography{mmm}
\bibliographystyle{unsrt}
\end{document}